\documentclass{article}

\usepackage{PRIMEarxiv}

\usepackage[utf8]{inputenc} 
\usepackage[T1]{fontenc}    
\usepackage{hyperref}       
\hypersetup{
        colorlinks = true,
        allcolors = magenta,
        breaklinks = true,
        linktocpage=true,
}
\usepackage{url}            
\usepackage{booktabs}       
\usepackage{amsfonts}       
\usepackage{nicefrac}       
\usepackage{microtype}      
\usepackage{lipsum}
\usepackage{fancyhdr}       
\usepackage{graphicx}       
\graphicspath{{media/}}     

\usepackage{multirow}%
\usepackage{amsmath,amssymb}%
\usepackage{amsthm}%
\usepackage{mathrsfs}%
\usepackage[title]{appendix}%
\usepackage{textcomp}%
\usepackage{manyfoot}%
\usepackage{booktabs}%
\usepackage{algorithm}%
\usepackage{algorithmicx}%
\usepackage{algpseudocode}%
\usepackage{listings}%
\usepackage{times}
\usepackage{epsfig}
\usepackage{subfiles} 
\usepackage{cite}

\usepackage{verbatim}
\usepackage{soul}
\usepackage{makecell}
\usepackage{pifont}
\newcommand{\cmark}{\ding{51}}%
\newcommand{\xmark}{\ding{55}}%
\newcommand{\comments}[1]{}

\title{JOSENet: A Joint Stream Embedding Network for Violence Detection in Surveillance Videos
}

\author{
  Pietro Nardelli and Danilo Comminiello \\
  Dept. of Information Engineering, Electronics and Telecommunications (DIET) \\
  Sapienza University of Rome \\
  Via Eudossiana 18 - 00184 Rome, Italy\\
}

\begin{document}
\maketitle

\begin{abstract}
The increasing proliferation of video surveillance cameras and the escalating demand for crime prevention have intensified interest in the task of violence detection within the research community. Compared to other action recognition tasks, violence detection in surveillance videos presents additional issues, such as the wide variety of real fight scenes. Unfortunately, existing datasets for violence detection are relatively small in comparison to those for other action recognition tasks. Moreover, surveillance footage often features different individuals in each video and varying backgrounds for each camera. In addition, fast detection of violent actions in real-life surveillance videos is crucial to prevent adverse outcomes, thus necessitating models that are optimized for reduced memory usage and computational costs. These challenges complicate the application of traditional action recognition methods.
To tackle all these issues, we introduce JOSENet, a novel self-supervised framework that provides outstanding performance for violence detection in surveillance videos. The proposed model processes two spatiotemporal video streams, namely RGB frames and optical flows, and incorporates a new regularized self-supervised learning approach for videos. JOSENet demonstrates improved performance compared to state-of-the-art methods, while utilizing only one-fourth of the frames per video segment and operating at a reduced frame rate. The source code is available at {\href{https://github.com/ispamm/JOSENet}{github.com/ispamm/JOSENet}}.
\end{abstract}


%
%
%
%
%
\section{Introduction}
\label{sec:intro}
Violence detection is a critical and challenging sub-task within the domain of human action recognition \cite{SuTCSVT2024, GuoTCSVT2024, FanTCSVT2024, ZhangTCSVT2017, akti_vision-based_2019}. Violent incidents, such as fights, can originate from various contexts and locations (e.g., burglaries, hate crimes), making early detection crucial for ensuring public safety \cite{perez-cctv-fight-2019}. However, there are limited tools available for the effective detection and prevention of violent actions \cite{ZhangTCSVT2022, sumon_violence_2020}. 
One of the most widely implemented measures to enhance public security is the deployment of Closed-Circuit Television (CCTV) video surveillance systems.
Despite their widespread use, CCTV systems necessitate extensive manual monitoring, which is susceptible to human fatigue and can compromise the speed and effectiveness of decision-making and crime prevention efforts \cite{xu_localization_2019, sernani_airtlab_2021}. 
An effective alternative solution for enhancing public safety is the development of deep learning methodologies for the automatic detection of violent actions \cite{islam_efficient_2021, sernani_airtlab_2021, sumon_violence_2020, ullah_intelligent_2021}.
However, detecting violent scenes in surveillance videos involves several challenges, including significant variations in actors and backgrounds across different videos, varying video lengths, and resource constraints imposed by real-time surveillance requirements. Additionally, the availability of labeled datasets suitable for effective supervised detection is limited, further complicating the task.
\begin{figure}[t]
    \centering
    \includegraphics[width=0.7\textwidth]{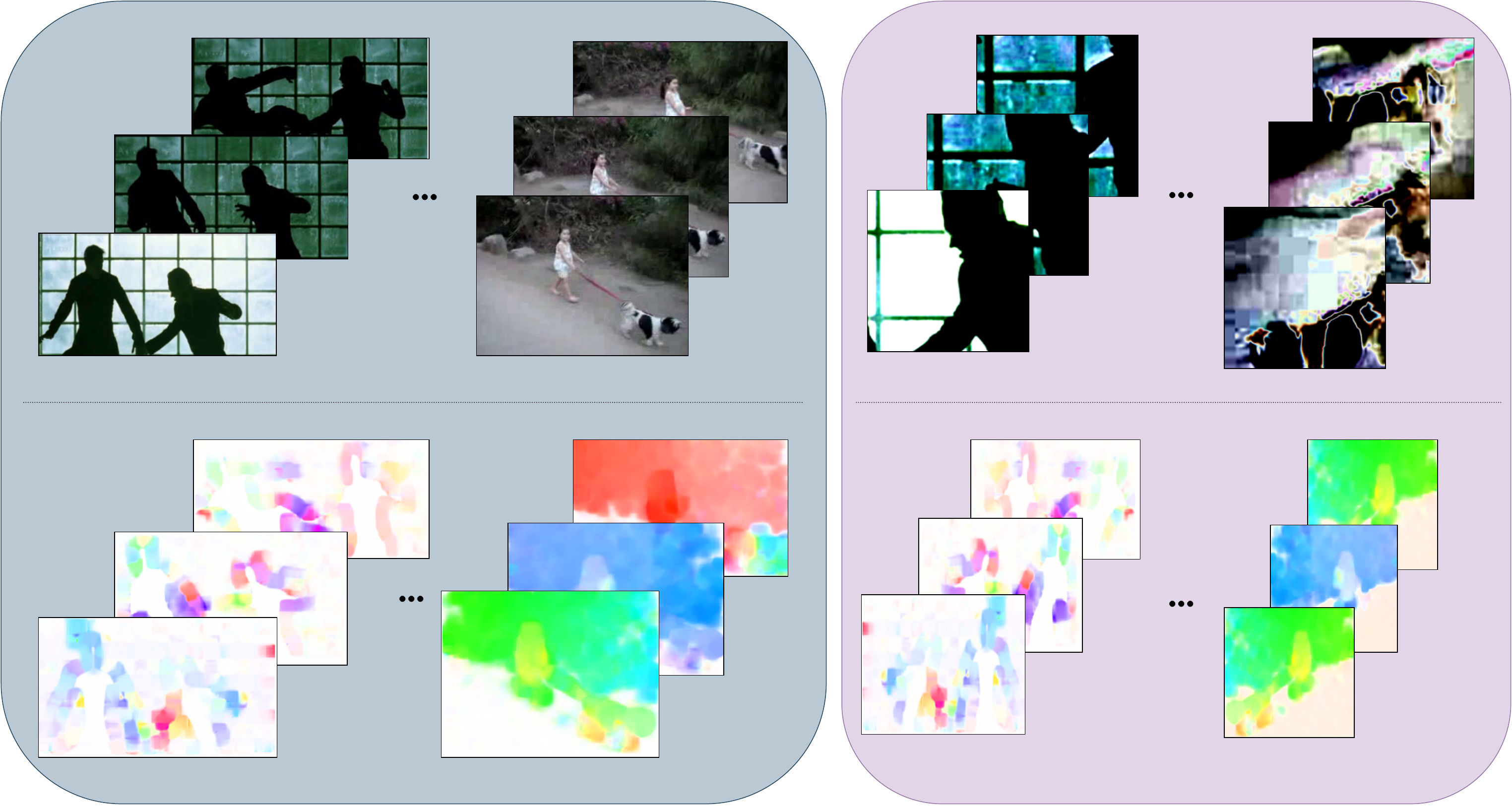}
    \caption{The proposed JOSENet architecture receives as input RGB and optical flow segments as a batch of size $N$, where each segment is made of $L$ frames. An example of the RGB (top-left) and optical flow segments (bottom-left) is shown here. Both RGB (top-right) and optical flow segments (bottom-right) are augmented. In particular, a strong random cropping strategy and some other augmentation techniques are applied to RGB frames while the optical flow segments are flipped horizontally.}
    \label{fig:augm}
\end{figure}

To address the aforementioned challenges in the violence detection task, this work presents JOSENet, an innovative joint stream embedding architecture. This architecture integrates an efficient multimodal video stream network with a novel self-supervised learning paradigm tailored for video streams. The proposed JOSENet receives two video streams, a spatial RGB flow and a temporal optical flow, as shown in Fig.~\ref{fig:augm} (left), which is a promising strategy in video understanding \cite{cheng_rwf-2000_2020, YuTIP2024}. 
Our method utilizes a minimal number of frames per segment and a low frame rate compared to state-of-the-art solutions, aiming to optimize the cost-benefit ratio from a production standpoint. However, this cost reduction may inadvertently impact performance accuracy. To mitigate this potential decrease in performance while preventing overfitting, we initialize the network with pretrained weights derived from self-supervised learning (SSL) approaches. These SSL methods generate effective representations without relying on human-annotated inputs, as shown in Fig.~\ref{fig:augm} (right). In addition to evaluating various state-of-the-art SSL methods compatible with JOSENet, we introduce a novel SSL algorithm specifically tailored for video streams. This algorithm is based on variance-invariance-covariance regularization (VICReg) \cite{bardes_vicreg_2022}, optimizing its performance for the unique characteristics of video data. 
To the best of our knowledge, this work represents the first instance of a VICReg-like video stream architecture. The proposed SSL method leverages the capabilities of VICReg, enabling it to i) scale effectively with the size of the data, ii) minimize memory requirements, and iii) prevent any collapse issues, distinguishing it from other existing methods.

The utilization of an SSL method renders JOSENet robust to the frequent scarcity of labeled data, a common issue in violence detection within real-life surveillance videos, thereby enhancing the model generalization capability \cite{yang_transfer_2020}. 
We prove the effectiveness of the proposed JOSENet framework over the most widely-used violence detection datasets and under various conditions, elucidating both the advantages and potential limitations of our method.
The experimental results indicate that the proposed JOSENet model effectively reduces both the number of frames per video and the frame rate, while achieving superior performance compared to existing SSL solutions.


The rest of the paper is organized as follows. Section~\ref{sec:relatedwork} presents the related work and the recent advances in the field. The proposed JOSENet framework is detailed in Section~\ref{sec:proposed_method}, including the novel primary and auxiliary models and the designed regularized self-supervised learning strategy. Extensive validation of the JOSENet framework is provided in Section~\ref{sec:results}, and further ablation studies are reported in Section~\ref{sec:ablation}. Finally, conclusion and future work are discussed in Section~\ref{sec:conclusion}.

%
%
%
%
%
\section{Related Work}
\label{sec:relatedwork}
\subsection{Violence Detection}
In the last years, the analysis of violent actions has become tractable thanks to deep neural networks.
An early work employed a VGG16 for optical flows \cite{mukherjee_fight_2017}.
A modified Xception convolutional neural network (CNN) is used in \cite{akti_vision-based_2019} together with a bidirectional long short-term memory (Bi-LSTM) model to learn the long-term dependency. 
Investigation on the use of bidirectional temporal encodings can be found in \cite{HansonECCV2019}. Bidimensional CNNs have been largely used for violent scene detection
. In particular in \cite{perez-cctv-fight-2019}, besides introducing the CCTV-Fight dataset, two 2D CNN VGG16 architectures were proposed, one for the spatial stream and another for the temporal stream. 
A framework with localization and recognition branches was proposed in \cite{xu_localization_2019}.
Alternative approaches are based on the 3D skeleton point clouds, e.g., extracted from videos via a pose detection module \cite{su_vedaldi_human_2020, GarciaCVIU2023, hachiuma_unified_2023}. 
In \cite{sumon_violence_2020}, several pretraining strategies were explored for detecting violence in videos. 
Efficient spatio-temporal architectures were proposed in \cite{islam_efficient_2021, KangACCESS2021}, leveraging ConvLSTM and fast pre-processing techniques.
%
\subsection{Self-Supervised Learning}
Self-supervised learning (SSL) aims to learn representations from unlabeled data, and build generalized models. 
The first SSL approaches were proposed for spatial context prediction \cite{doersch_unsupervised_2016} and Jigsaw puzzle solution \cite{noroozi_unsupervised_2017}.
The contrastive learning (CL) approach \cite{dosovitskiy_discriminative_2015} discriminates between a set of augmented labels. The contrastive predictive coding \cite{infonce_2019} extracts useful representations from high-dimensional data. In \cite{haresamudram_contrastive_2020}, CL was applied to human activity recognition.
While CL obtained competitive performance with respect to supervised representation \cite{chen_simple_2020, he_momentum_2020}, it does not scale well with the dimension of the data, and it tends to require large memory demands. Regularized methods solve these problems \cite{grill_bootstrap_2020, caron_unsupervised_2021}. 
In particular, the Barlow twins method \cite{zbontar_barlow_2021} naturally avoids collapse by the measure of cross-correlation matrix between the two outputs of a Siamese neural network, fed with distorted samples. 
Inspired by \cite{zbontar_barlow_2021}, the variance-invariance-covariance regularization (VICReg) \cite{bardes_vicreg_2022} has been proposed for multimodal data, showing good scaling ability and limited memory demand. This paper introduces a novel variant of VICReg in the JOSENet framework, specifically proposed for video streams.
%
\subsection{Self-Supervised Learning for Video}
Several SSL techniques were specifically proposed for video streams, including shuffle and learn, odd-one-out(O3D), and methods based on Siamese networks \cite{misra_shuffle_2016, fernando_self-supervised_2017, lee_unsupervised_2017}. 
One of the first works acting on a two-stream architecture was presented in \cite{taha_two_2018}.
Understanding the forward or backward video playing direction was the focus of the arrow of time (AoT) \cite{wei_learning_2018}.
Spatiotemporal 3D CNNs were introduced in SSL as the space-time cubic puzzle \cite{kim_self-supervised_2019}, a context-based method that requires arranging randomly permuted spatiotemporal crops of a video using a 4-tower Siamese network.
In \cite{wang_self-supervised_2019}, a model was defined to predict numerical labels based on statistical concepts. 
A method based on contrastive predictive coding for video representation learning was developed in \cite{lorre_temporal_2020}. Dense predictive coding was proposed for learning spatiotemporal video embeddings \cite{han_video_2019}, without relying on additional modalities, such as the optical flow. Differently, CoCLR \cite{co_clr_2020} exploits complementary data (i.e., optical flow) as additional positive samples in a new co-training regime. 
The pretext-contrastive learning (PCL) \cite{tao_pretext-contrastive_2021} is a joint optimization framework for both CT and pretext tasks (e.g., jigsaw or rotation). 
The only work that uses SSL for the violence detection task introduces an iterative learning framework based on both weakly and self-supervised paradigms, where two experts feed data to each other where the SSL expert is a C3D network \cite{degardin_human_2020}. The expert classifier involved in our JOSENet framework can be seen as a novel advanced version of the C3D network.
Also, in \cite{seo_self-supervised_2022}, an SSL approach is adopted to pretrain a module for selecting informative frames for abnormal action recognition. Our aim instead, is to obtain a better-performing network without using additional modules that could slow down the inference speed, which is critical for violence detection.
\begin{figure}[t]
\centering
\includegraphics[width=0.65\textwidth]{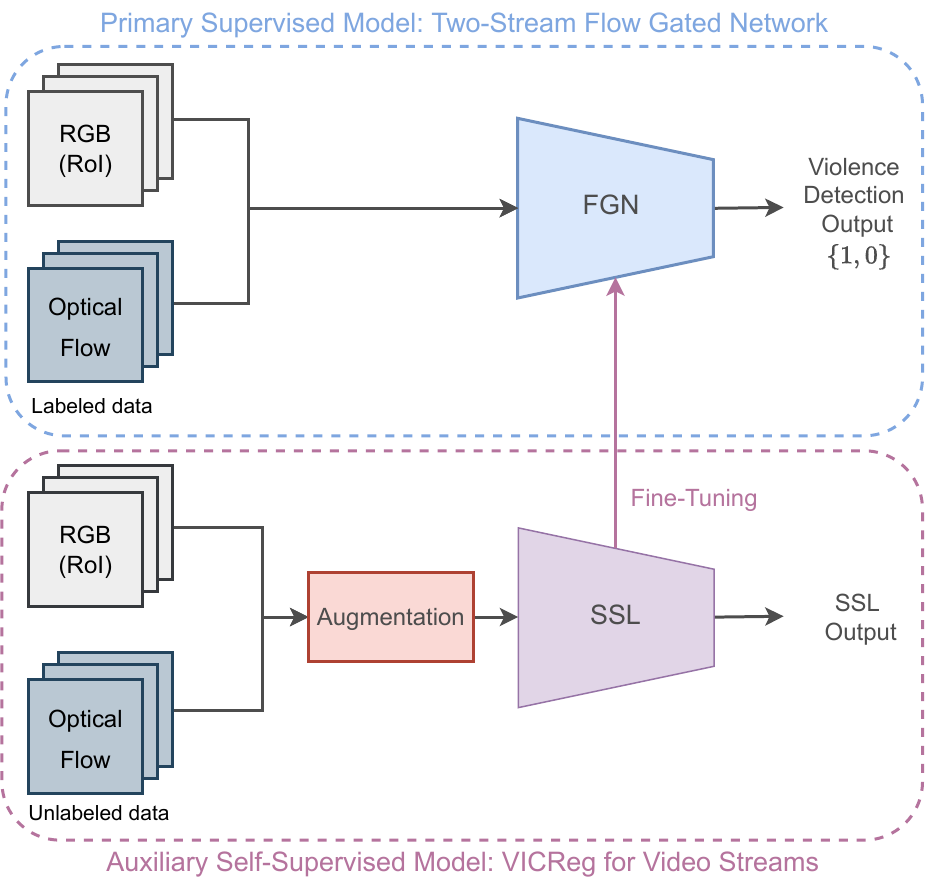}
\caption{The proposed JOSENet framework. The primary target model (top) is tackled by using a novel efficient flow gated network (FGN) which produces binary classification (1 if violence is detected, 0 otherwise) given optical flow and RGB segments. The FGN is pretrained by using a novel two-stream SSL method (bottom) that aims to solve an auxiliary task with unlabeled input data.}
\label{fig:framework}
\end{figure}

%
%
%
%
%
\section{The Proposed JOSENet Framework}
\label{sec:proposed_method}
%
\subsection{Framework Description}
The proposed JOSENet framework for violence detection is basically composed of two parts, as depicted in Fig.~\ref{fig:framework}: a primary model and an auxiliary SSL model. 

The primary part comprises a two-stream architecture involving both spatial and temporal flows. This choice usually guarantees significant performance in detection tasks in video \cite{YuTIP2024}. In particular, JOSENet involves a two-stream flow gated network (FGN), which performs violent action detection using labeled data for both the spatial and the temporal flows. To reduce both memory and computational costs, we use a lightweight setting for the model in terms of the number of frames per video. 

The JOSENet framework benefits from an auxiliary model to avoid any performance loss. This auxiliary network implements a novel SSL method receiving unlabeled data as input. The weights of the auxiliary model are optimized by minimizing the VICReg loss, and subsequently employed in the pretraining of the primary supervised model. Following the pretraining phase, a fine-tuning strategy is utilized to refine and tailor the pretrained weights to the specific requirements of the primary model. The auxiliary SSL network allows JOSENet to achieve the best trade-off between performance and employed resources.
In the following, we focus in detail on the two models of the proposed framework.
%
\subsection{The Primary Model: An Efficient Two-Stream Flow Gated Network}
The primary target model is based on a two-stream FGN. The choice of a two-stream architecture is motivated by the benefits brought by multimodal video architectures and by the excellent performance that the FGN achieved on the RWF-2000 dataset \cite{cheng_rwf-2000_2020}. 
This primary network consists of three modules: a spatial block, a temporal block, and a merging block.

%
\textbf{Spatial Block.} The spatial RGB module receives as input consecutive frames that are cropped to extract the region of interest (ROI). The ROI aims to reduce the amount of input video data, making the network focus only on the area with larger motion intensity. 
The computation of the ROI involves normalization and a subtraction of the mean of each optical flow frame for denoising purposes. 
Given the normalized and denoised optical flow frame $S_i$, the magnitude can be computed as $(S_{i,x}^2+S_{i,y}^2)^{\frac{1}{2}}$ where $S_{i,j}$ represents the $j$-th component of the $i$-th frame. 
The sum of the magnitudes of each frame produces a $224 \times 224$ motion intensity map, on which the mean is computed and used as a threshold to additionally filter out the noise (i.e., zeroing out the motion intensity map values if less than this threshold).  
To obtain the center of the ROI based on the motion intensity map, a probability density function along the two dimensions $x, y$ of the motion intensity map is used. Ten different candidates are selected to be the center of the ROI, and are extracted randomly from this probability density function. The final value of the center $(c_x, c_y)$ is obtained by the average of these 10 points for better robustness.
The ROI is extracted by a patch of size $112 \times 112$ from the RGB frames based on $(c_x, c_y)$, thus a cubic interpolation is applied to reconstruct $N$ frames with size $224 \times 224$.
Once processed, the output of the RGB block passes through a ReLU activation function. 
The resulting input dimension of the RGB block is $3 \times N \times 224 \times 224$, where the first dimension represents the RGB channels of the video segment. 

%
\textbf{Temporal Block.} 
The temporal block receives as input the optical flow of the sequence of frames. The flow is computed by using the Gunnar Farneback's algorithm, as in \cite{cheng_rwf-2000_2020}. 
For each RGB segment $(F_0, F_1, \dots, F_s)$ of $N$ frames of size $224 \times 224$, an optical flow frame is computed from each couple $(F_{i-1}, F_{i})$. The resulting dimension of the flow block is $2 \times N \times 224 \times 224$.
A sigmoid activation function at the end of this branch scales the output for the RGB embeddings. 

%
\textbf{Merging Block.} 
The last FGN module defines the fusion strategy that manages both the RGB and the flow streams. In particular, the output of the RGB block and the flow block are multiplied together and processed by a temporal max pooling. 
This is a self-learned pooling strategy that utilizes the flow block as a gate, aiming to decide which information from the RGB block should be maintained or dropped.
Finally, the fully-connected layers generate the output for that input. 

%
\subsection{Computational Efficiency of the Primary Model}
%
\textbf{Computational Enhancement} 
Since violence detection is a real-time application, it is necessary to reduce the computational cost as much as possible, thus finding more efficient ways to produce inference \cite{xu_localization_2019}, while deploying the model with reduced memory usage and frame rate during inference. To maintain the cost as low as possible, it is necessary to first deal with the number of frames $N$ in the input video segment while, at the same time, the network should be able to learn the correct features by using a correct window size $T_{\text{tf}}$ (temporal footprint). 
The vanilla FGN \cite{cheng_rwf-2000_2020} uses $N = 64$ frames with $N_{fps} = 12.8$ frames per second (FPS). Instead, to speed up the inference, we use $N = 16$ which is a common value for the most used action recognition architectures such as R(2+1)D \cite{tran_closer_2018}, C3D \cite{tran_learning_2015}, I3D \cite{carreira_quo_2018} and P3D \cite{qiu_learning_2017}.

In action recognition, it is known that a very short window can lead to perfect recognition of most activities \cite{banos_window_2014} while at the same time, the classification performance generally increases by using a very high frame rate.
However, as pointed out in \cite{harjanto_investigating_2016}, action recognition methods do not always obtain their best performance at higher frame rates, but the best results are achieved by each method at different frame rates. In this way, we treat $N_{fps}$ like a hyperparameter that we aim to reduce for practical reasons. Thus, we find that a value of $N_{fps} = 7.5$ is an optimal trade-off between computational cost and performances, obtaining as a result a temporal footprint of $T_{\text{tf}} = \frac{N}{N_{fps}}=2.13$s. This choice can be considered appropriate for the broad category of violent actions. Further generalization performance can be found in Section \ref{sec:results}.

%
\textbf{Efficient Implementation.} 
To reduce the segment length $N$ we modify the $2 \times 2 \times 2$ max-pooling layers of the merging block into a $1 \times 2 \times 2$ (i.e., reducing by 1 the temporal dimension) so that the output dimension for that block is unchanged.
In addition, to approximately halve the memory requirements while speeding up arithmetic, we use mixed-precision \cite{narang_micikevicius_mixed_2018}. 
With the aim of reducing the internal covariance shift, a 3D batch-norm layer is applied after each activation function of all the blocks of our architecture, except for the fully connected (FC) layers. 
Lastly, to avoid overfitting, spatial dropout with $p = 0.2$ is applied after each batch normalization layer of the first two 3D convolutional blocks, in both RGB and optical flow blocks. 

The proposed model achieves a significant reduction of computational complexity compared to the original network in \cite{cheng_rwf-2000_2020}. Specifically, our model required only 4.432G multiply-accumulate operations (MACs), whereas the original architecture demanded 33.106G MACs, indicating a \textbf{7-fold reduction in computational load}. In addition, through a 4x smaller segment size and a reduced temporal footprint, we are able to significantly \textbf{reduce the memory requirements by 75\%} and achieve a \textbf{two-fold increase in signaling alarm speed in real-life scenarios}. This result highlights the potential of our framework for efficient and effective neural network design. One might view these findings as addressing an inefficiency rather than introducing a new algorithm — yet, this perspective does not diminish their significance.

%
\subsection{The Auxiliary Model: VICReg for Joint Video Stream Architectures}
\label{subs:auxiliary}
%
\textbf{Self-Supervised Pretraining.} 
The original FGN \cite{cheng_rwf-2000_2020} does not involve any pretraining. While on some occasions the pretraining may not increase the performances on classification metrics, it is demonstrated that it can improve the model robustness \cite{hendrycks_using_2019}. Furthermore, using data from multiple sources helps generalize the model better for a two-stream architecture \cite{perez-cctv-fight-2019}.
In this work instead, we investigate the use of self-supervised pretraining as opposed to fine-tuning approaches 1) to compensate for the performance loss due to the resource limitation of the primary network, 2) to deal with unlabeled data typical of real-world surveillance video applications, and 3) to improve the generalization performance by avoiding a bias toward the source labels on the source task \cite{yang_transfer_2020}. 
We implemented four different SSL techniques: odd-one-out (O3D) \cite{fernando_self-supervised_2017}, arrow of time (AoT) \cite{wei_learning_2018}, space-time cubic puzzle (STCP) \cite{kim_self-supervised_2019}, and VICReg \cite{bardes_vicreg_2022}. 
Clearly, all the above SSL techniques are fairly adapted to our FGN. 

\begin{figure*}[t]
    \centering
    \includegraphics[width=\textwidth]{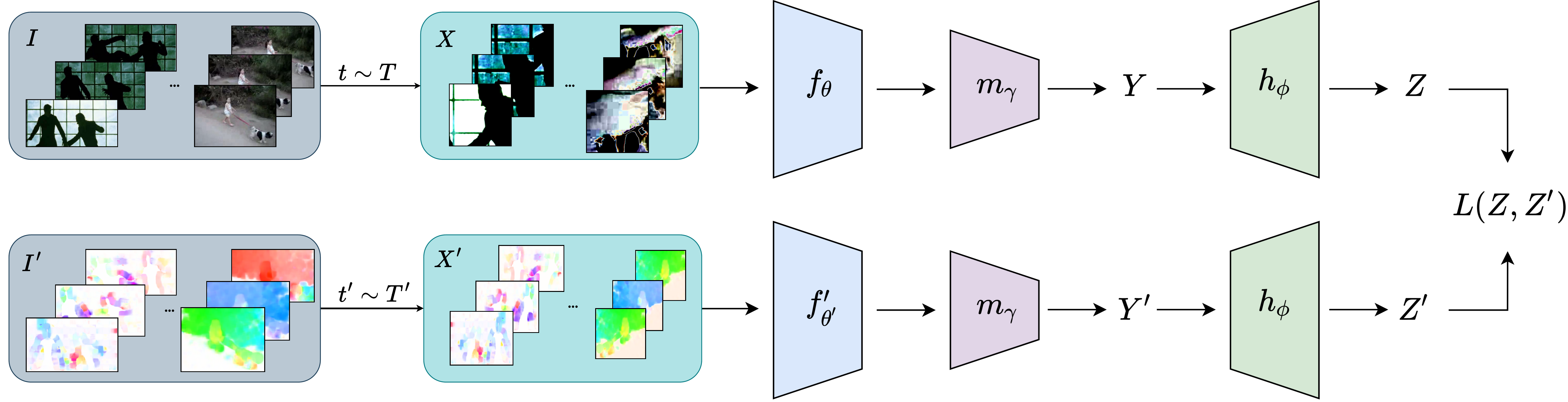}
    \caption{The proposed VICReg solution for the auxiliary model of the JOSENet framework. $I$ and $I'$ are respectively a batch of RGB and flow segments that are transformed through data augmentation into two different views $X$ and $X'$. In particular, a strong random cropping strategy and some other augmentation techniques are applied to RGB frames while the flow frames are only flipped horizontally. The RGB branch is represented by $f_\theta$, the optical flow branch is $f'_{\theta'}$, $m_\gamma$ is the merging block without the temporal max pooling and finally, the $h_\phi$ is the expander as in the VICReg original implementation. The VICReg loss function $L(Z, Z')$ is computed on the embeddings $Z$ and $Z'$.}
\label{fig:my_vic_reg}
\end{figure*}

%
\textbf{A novel VICReg for JOSENet.} 
In our framework, we focus on the variance-invariance-covariance regularization (VICReg) \cite{bardes_vicreg_2022}, an SSL method for joint embedding architectures that preserves the information content of the embeddings, while not demanding large memory requirements, contrastive samples nor memory banks, unlike what happens in contrastive methods \cite{jing_understanding_2022}.
To the best of our knowledge, for the first time VICReg is applied to video streams
. The proposed VICReg solution for JOSENet relies on the joint information of the augmented RGB and flow batches and it is depicted in Fig.~\ref{fig:my_vic_reg}. 

Let us consider an RGB batch $I$ and an optical flow batch $I’$ both related to the same input segment. Two augmented views of these batch $X$ and $X'$ can be produced by using random transformations $t$ and $t'$ sampled from a distribution $T$. The augmented batches are fed into two different encoders $f_{\theta}$ and $f'_{\theta'}$. The two branches do not have the same architectures and do not share the same weights. The output of these branches is fed as input into two Siamese merging blocks $m_{\gamma}$ that have the same architecture as the merging block but without the temporal max pooling. Indeed, the removal of this pooling seems to be beneficial thanks to the increase of the expander input dimensionality. The output representations $Y$ and $Y'$ are fed as input into two Siamese expanders $h_{\phi}$ which produces batches of embeddings $Z = h_{\phi}(Y)$ and $Z' = h_{\phi}(Y')$ of $n$ vectors of dimension $d$. 
By utilizing the Siamese merging block during the self-supervised phase, we are able to generate representations of the input data by leveraging a significant portion of the FGN architecture. As a result of this approach, we are able to obtain highly informative and useful feature representations of the input data, which can subsequently be utilized for the primary task. This particular solution would be infeasible with other SSL techniques and can be designed only by using a method that can handle multimodality, such as VICReg. Indeed, among the several configurations tested, we will demonstrate that this is the best setup for pretraining our two-stream FGN.
We leverage the basic idea of VICReg to use a loss function with three different terms. The variance regularization term $v(Z)$ is computed along the batch dimension as the standard deviation of the embeddings and aims to prevent a complete collapse. 
The covariance regularization term $c (Z)$ encourages the network to decorrelate the dimensions of the embeddings so that similar information is not encoded. This kind of decorrelation at the embedding level leads to a decorrelation at the representation level as well. Our approach entails to diminish the disparities between the two data modalities through the minimization of the invariance term $s (Z, Z')$ which is the mean-squared Euclidean distance between each pair of embedding vectors. The overall loss function $L (Z, Z')$ is a weighted average of these terms:
\begin{equation}
    L (Z, Z') = \lambda s(Z, Z') + \mu [v(Z) + v(Z')] + \nu [c(Z) + c(Z')]
    \label{eq:vicregloss}
\end{equation}
where $\lambda, \mu$ and $\nu$ are hyperparameters (specifically, $\lambda = \mu = 25$ and $\nu = 1$ works best in most of the contexts).
Since VICReg is an information maximization method, it does not require the use of techniques generally used in contrastive methods. Moreover, although it was proposed for a Siamese network \cite{bardes_vicreg_2022}, one of its greatest advantages is that the two branches could also not share the same parameters, architectures, and more importantly input modality.

%
\textbf{Configuration.} 
In \cite{bardes_vicreg_2022}, the input size of the expanders $h_\phi$ is set to 2048. In our case, using a merging block $m_\gamma$ with an unchanged structure would produce an input of size 128. Since it is paramount to 
include most of the two-stream architecture without reducing too much the expander dimensionality, we remove the temporal max-pooling in the merging block. In this way, the expander input dimensionality grows from 128 to 1024, finding a good trade-off between the two constraints. Thus, the output representations $Y$ and $Y'$ have dimension 1024 and are fed as input into the two Siamese expanders $h_{\phi}$. The expanders have the same structure as the original method: 3 FC layers of size 8192, where the first two layers utilize batch normalization and ReLU. 

%
%
%
%
%
\section{Experimental Setup}
\label{sec:setup}
In this Section, we briefly discuss all the settings that make JOSENet an effective solution for violence detection tasks.

%
\subsection{Datasets} 
During the last ten years, the interest in violence detection systems has significantly grown and several datasets have been developed \cite{perez-cctv-fight-2019, akti_vision-based_2019, sernani_airtlab_2021}. 
To train and validate the primary model of the JOSENet framework for supervised learning we use the RWF-2000 dataset \cite{cheng_rwf-2000_2020}, involving 2,000 heterogeneous videos, 5 seconds long, and captured at $N_{fps} = 30$ FPS by real-world surveillance cameras. 
In line with the existing literature, we also use HMDB51 \cite{hmdb51_2011} (51 classes spread over 6,766 clips) and UCF101 \cite{soomro_ucf101_2012} (nearly twice the size of HDMB51) datasets.
Although a larger dataset can be beneficial in most cases, SSL techniques generally outperform transfer learning when the amount of pretraining is small \cite{yang_transfer_2020}. Moreover, since a strong domain similarity can be very useful \cite{yang_transfer_2020}, we use the UCF-Crime \cite{sultani-ucf-crime-2019} as an additional dataset.
We maintain the default train-test split in all the datasets used in both target and auxiliary models. 

%
\subsection{Preprocessing Methods}
We use the same frame resolution as in \cite{cheng_rwf-2000_2020} ($224 \times 224$) with each video segment generated with $N_{fps} = 7.5$ FPS in a sliding window manner \cite{banos_window_2014}. 
For the AoT, a total of $32+1$ RGB frames are used to generate 32 optical flow frames that are processed by the AoT network. For the STCP, we sample $N = 64$ frames, removing the temporal jittering in the STCP auxiliary model. 
To prevent a slow convergence during training, we avoid the upside-down flip as suggested in \cite{kim_self-supervised_2019}. Thus, the total number of classes for the STCP approach is $c = 24$. Lastly, both O3D and VICReg receive $16$ input frames as the target model.

%
\subsection{Zoom Crop: The Proposed Data Augmentation Strategy}
%
\textbf{Data Augmentation for the Primary Model.}
For the primary model, we apply color jitter to RGB frames and random flip to both RGB and flow frames. The color jitter randomly changes the brightness, contrast, saturation, and hue of images within a range of $[-0.2,0.2]$. Both color jitter and random flip are applied with a probability of 50\%. To maintain coherence between optical flow and RGB frames, the random flip is applied to both of them. After the data augmentation (in both the primary and auxiliary models), a standardization is applied per input vector, as in \cite{cheng_rwf-2000_2020}. 
In order to have a fair comparison, we use the best augmentation strategy for each SSL approach. 

%
\textbf{The Zoom Crop Strategy for the Auxiliary Model.}
For the proposed VICReg SSL approach of the JOSENet auxiliary model, we use a stronger random cropping strategy in the RGB augmentation pipeline compared to the original approach \cite{bardes_vicreg_2022}. The strategy involves a scaling factor within the range $[0.08, 0.1]$, which is confirmed to be more efficient from our tests. We call this augmentation technique the ``zoom crop" strategy.  
The goal of the proposed strategy is to further enhance the exploitation of local features produced by a cropped view of the RGB frames.

Regarding the flow segments, a horizontal flip is applied with a probability of 50\%. This approach seems to be coherent with the features learned by the primary model. In fact, the RGB branch of the FGN receives as input a cropped version of the RGB segment (ROI), while the flow segment remains unchanged. 
A visualization of the VICReg video input for the JOSENet framework can be seen in Fig.~\ref{fig:my_vic_reg}. 

%
\subsection{Metrics}
Similarly to \cite{sernani_airtlab_2021}, we evaluate the performance by the following metrics: the accuracy, F1-score, true negative rate (TNR), true positive rate (TPR), and the area under the curve (AUC) to understand the model diagnostic capability in identifying violent videos. We enhance model performance by optimizing accuracy, the primary evaluation metric in most violence detection methodologies. However, we acknowledge the ethical responsibilities associated with violence detection applications. Therefore, we prioritized metrics such as TNR to minimize the false positive rate, thereby reducing the risk of innocent individuals being wrongly accused or flagged as potential threats.

Unless otherwise stated, for the experiments we used an NVIDIA RTX5000 Quadro GPU with 16GB GDDR6 and 8 virtual cores by Intel Xeon 4215 CPU with a frequency of 3.2 GHz. 

%
\subsection{Definition and Results of the Baseline Model Without Pretraining}
\label{sec:experiments_baseline}
The initial experiments are designed to identify the optimal baseline model without self-supervised pretraining. Our primary focus is on hyperparameter tuning, through which we determined the most effective parameters for the network (see Table~\ref{tab:hyperparameters}). 
More specifically, we train the network on 32 batches for a total of 30 epochs together with an early stopping procedure with a patience of 15 epochs.
The number of frames for each segment is $N = 16$, sampled at $N_{fps} = 7.5$ FPS. We use a $p = 0.2$ dropout probability for the classification block.
A binary cross-entropy loss is employed with a stochastic gradient descent (SGD) optimizer (momentum 0.9 and 1e-6 weight decay), as it is a state-of-the-art choice for violent detection \cite{cheng_rwf-2000_2020}. A cosine annealing scheduler is implemented, which starts from the initial learning rate value of 0.01, and decreases for 30 epochs to reach a minimum of 0.001.
\begin{table}[t]
\caption{The optimal hyperparameters used for all the primary models.}
\label{tab:hyperparameters}
\begin{center}
\begin{tabular}{ |c|c|c| } 
\hline
Parameter & Value \\
\hline
\texttt{clip\_frames} & 16 \\ 
\texttt{fps} & 7.5 \\ 
\texttt{batch\_size} & 32 \\ 
\texttt{dropout} & 0.2 \\ 
\texttt{epochs} & 30 \\ 
\texttt{patience} & 15 \\ 
\texttt{learning\_rate} & 0.01 \\ 
\texttt{momemtum} & 0.9 \\ 
\texttt{weight\_decay} & 1e-6 \\
\texttt{T\_max} & 30 \\
\texttt{eta\_min} & 0.001 \\
\hline
\end{tabular}
\end{center}
\end{table}

With these settings, we obtain 84.25\% accuracy. To avoid overfitting at this stage, we add spatial dropout with a probability of 0.2. 
In this case, we obtain 85.87\% accuracy, thus we increase it by +1.62\%, F1-Score 85.87\%, 85.5\% TNR, 86.25\% TPR, and 0.924 AUC, as also summarized in Table~\ref{tab:baseline_results}. We use this model as a ``baseline" for comparisons.

\begin{table}
\centering
\caption{Results obtained by the defined baseline model for the RWF-2000 dataset.}
\label{tab:baseline_results}
\begin{tabular}{ |c|c|c|c|c| } 
\hline
Accuracy [\%] & F1 [\%] & TNR [\%] & TPR [\%] & AUC \\
\hline
85.87 & 85.87 & 85.5 & 86.25 & 0.924 \\
\hline
\end{tabular}
\end{table}

%
%
%
%
%
\section{Experimental Results}
\label{sec:results}

%
Now we evaluate the whole JOSENet framework, where the primary model benefits from the embeddings obtained by the SSL pretraining. 
For each method in the comparisons, we pretrain on three different datasets: HMDB51, UCF101, and UCF-Crime. 
During pretraining we use the same hyperparameters of the primary model with the difference in weight decay value of 1e-6 while epochs and batch size vary based on the technique used. 
In each experiment, we maintain the maximum number of iterations of the cosine annealing scheduler equal to the number of epochs.
While for VICReg we use the custom loss described by {eq.~\ref{eq:vicregloss}}, in all the other techniques a cross-entropy loss is applied. 
A fine-tuning strategy is applied by training the primary model on the target task. 

\begin{table*}
\centering
\caption{Results obtained by 
pretraining 
our FGN using 
the non-regularized SSL methods 
and fine-tuning it on the target task
. Each SSL method is tested with a different dataset 
for pretraining
.}
\vspace{0.1cm}
\label{tab:non-regular_ssl_results}
\begin{tabular}{ |c|c|c|c|c|c|c| } 
\hline
SSL Approach & Dataset & Accuracy [\%] & F1 [\%] & TNR [\%] & TPR [\%] & AUC \\
\hline
O3D & HMDB51 & 82.62 & 82.62 & 84 & 81.25 & 0.897 \\
O3D & UCF101 & 82.75 & 82.74 & 82.25 & 83.25 & 0.888 \\
O3D & UCF-Crime & 83.37 & 83.37  & 83.75 & 83 & 0.901 \\
\hline
AoT & HMDB51 & 84.75 & 84.74 & 86.25 & 83.25 & 0.911 \\
AoT & UCF101 & 84.5 & 84.49 & 82.75  & 86.25 & 0.911 \\
AoT & UCF-Crime & 84 & 83.97 & 80 & 88& 0.900 \\
\hline
STCP & HMDB51 & - & -  & -  & - & - \\
\textbf{STCP} & \textbf{UCF101} & \textbf{86.25} &  \textbf{86.25} & 85.25 & \textbf{87.25} & \textbf{0.913}\\
STCP & UCF-Crime & 85.25 & 85.23  & \textbf{88.75} & 81.75 & 0.912\\
\hline
\end{tabular}
\end{table*}

%
%
\subsection{JOSENet with Non-Regularized SSL Approaches} 
We evaluate the proposed JOSENet framework using non-regularized SSL methods in the auxiliary model. In particular, we consider O3D \cite{fernando_self-supervised_2017}, AoT \cite{wei_learning_2018} and STCP \cite{kim_self-supervised_2019}. Results are shown in Table~\ref{tab:non-regular_ssl_results}. 

As expected, the O3D technique does not provide useful embeddings for the primary model. In fact, the results are all worse than the baseline model by a wide margin. It is useful to note that the results are better in UCF-Crime, while HMDB51 produces the worst pretraining weights. 
For what concerns the AoT method, the results are unsatisfactory, probably because this technique (like the O3D one) is strongly dependent on the architecture chosen (in both cases they rely on 2D CNNs). It is interesting to see that the HMDB51 and UCF101 obtain similar results while UCF-Crime produces the worst performance. This may be due to a trivial learning problem due to low-level cues. This is also confirmed by the high accuracy obtained on the auxiliary task itself. 
For STCP some improvements are achieved by using the UCF101 dataset.
It is important to point out that we obtained a slight improvement compared to the baseline model and higher results compared with the other SSL pretraining. Differently from the previous methods, the STCP is built to be used on 3D CNNs, confirming that the architecture similarity plays an important role in the choice of the correct non-regularized SSL technique used.


%
\begin{table*}[t]
\centering
\caption{Results obtained by our proposed VICReg method on a random 15\% subset of the datasets UCF101 (R-UCF101) and UCF-Crime (R-UCF-Crime). In particular, R-UCF-Crime surpasses by a wide margin R-UCF101 on TPR and more importantly on F1-score. The final results obtained with a pretraining on the entire UCF-Crime dataset are shown in the last row.}
\vspace{0.1cm}
\label{tab:dataset_vicreg_results}
\begin{tabular}{ |c|c|c|c|c|c| } 
\hline
Dataset & Accuracy [\%] & F1 [\%] & TNR [\%] & TPR [\%] & AUC \\
\hline
HMDB51& 85.37 & 85.37 & 84 & 86.75 & 0.924\\
R-UCF101 & 86.37 & 86.37 & 85.75 & 87 & 0.922\\
R-UCF-Crime & \textbf{86.75} & \textbf{86.74} & 84.5 & \textbf{89} & 0.918 \\
\hline
UCF-Crime & 86.5 & {86.5} & \textbf{88} & {85} & \textbf{0.924}\\
\hline
\end{tabular}
\end{table*}

%
\subsection{JOSENet with Pretraining Dataset Selection for VICReg}
Many different experiments are carried out to define the best configuration for the regularized SSL method VICReg (e.g., Siamese RGB/Flow, augmentation tuning) by using the HMDB51 dataset (see Section~\ref{sec:ablation}).
We focus the attention on an approach without shared weights by pretraining simultaneously both RGB and optical flow branches. 
We assume that the best configuration involves a zoom crop strategy without considering the temporal max-pooling in the merging block.
To validate this hypothesis and understand the best dataset for pertaining we test on the remaining two datasets: UCF101 and UCF-Crime. To provide a fair comparison between HDMB51 and the remaining ones, while maintaining a feasible training time (i.e., each pretraining on the HMDB51 dataset lasts around 12 hours), in this phase, we pretrain on a subset of UCF101 and UCF-Crime, denoted as R-UCF101 and R-UCF-Crime respectively. The ``R" signifies ``Reduced", indicating that we randomly sample 15\% of each dataset, using 16 batches for training.
The results are shown in Table~\ref{tab:dataset_vicreg_results}.
As we can see, VICReg pretrained on R-UCF101 outperforms the best model obtained so far. Compared to the baseline, we show an increase in accuracy (+0.5\%), F1-score (+0.5\%), TNR (+0.25\%), and TPR (+0.75\%). 
Similarly, the model pretrained on R-UCF-Crime and tested on the target task shows a huge increase in performance in terms of F1-score (+0.87 \%) and TPR (+2.75 \%). This result underscores the importance of the quality of the dataset for the auxiliary model. 
Moreover, as expected, the dataset domain similarity between auxiliary and target tasks seems to be beneficial, resulting in an important factor to be considered when choosing the pretraining dataset.


%
\subsection{Final Results for JOSENet} 
\begin{figure*}[t]
\centering
\includegraphics[width=0.35\textwidth]{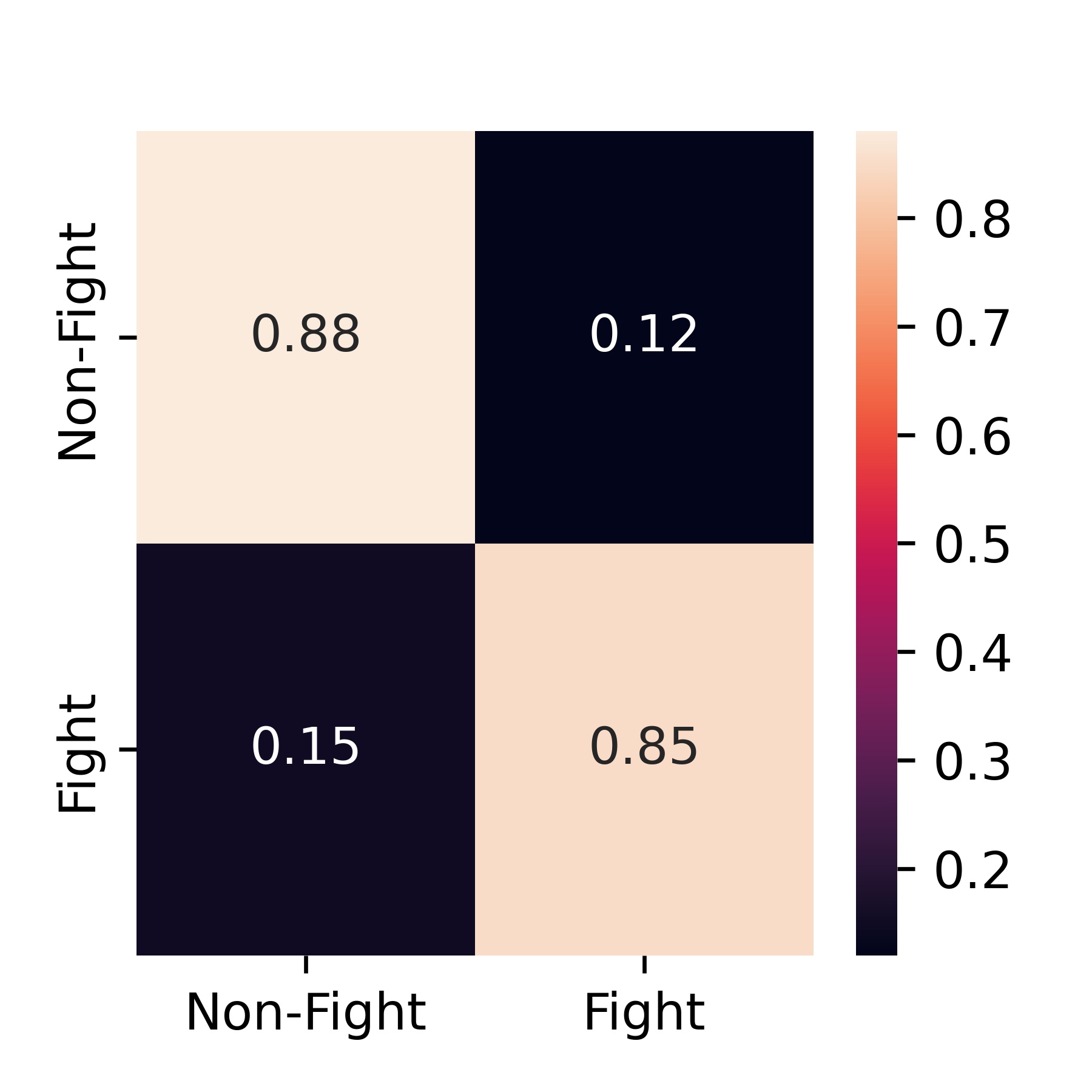}
\includegraphics[width=0.53\textwidth]{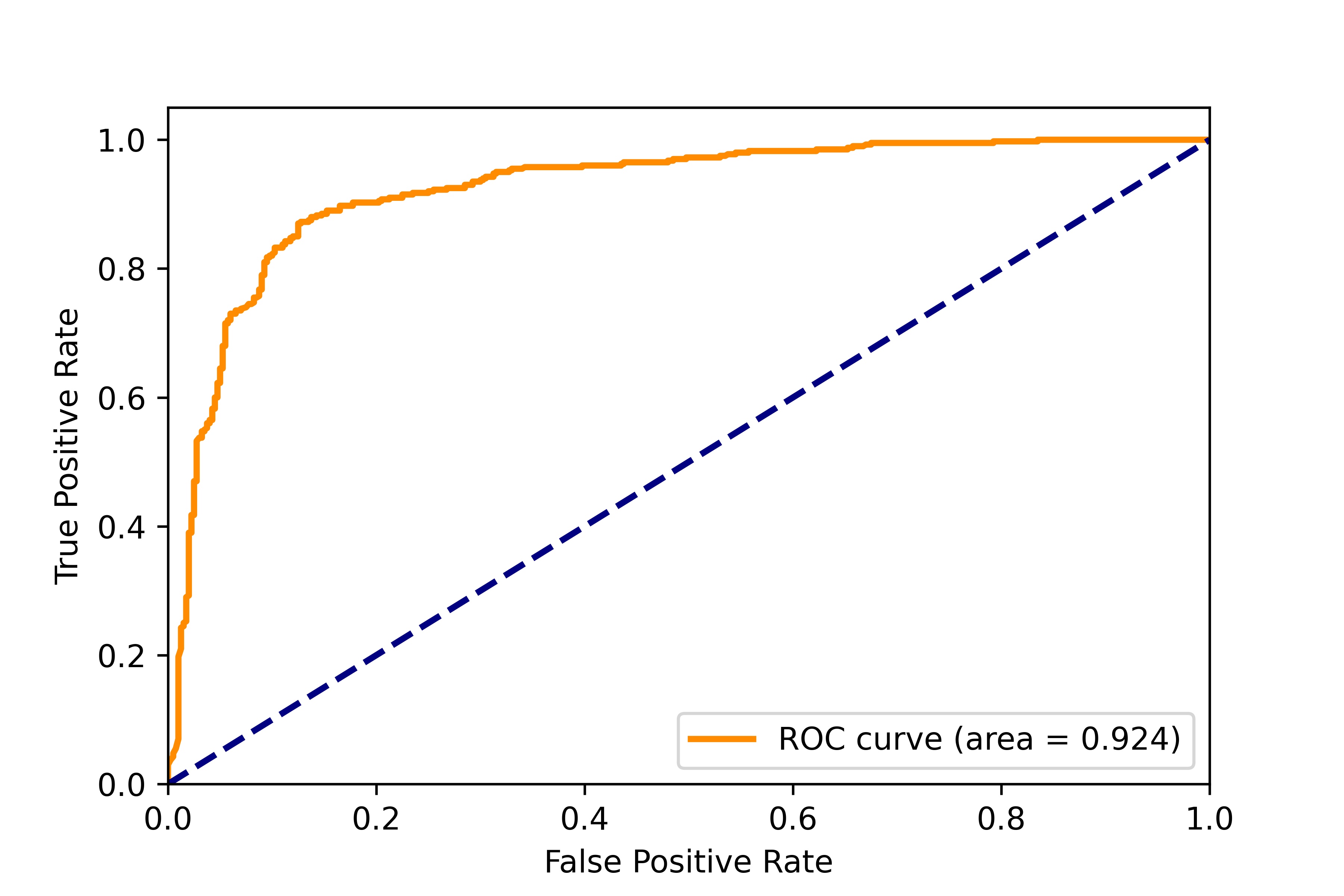}
\caption{Normalized confusion matrix (left) and ROC curve (right) obtained by evaluating our best model on the RWF-2000 validation set by pretraining via our VICReg proposed method with 64 batch size on the entire UCF-Crime dataset. The rows and columns of the confusion matrix represent respectively the predicted and target labels.}
\label{fig:roc_and_confusion}
\end{figure*}

To obtain our best performances and to further validate our solution, we train our best configuration on the entire UCF-Crime dataset using an NVIDIA Tesla V100S GPU with 32GB GPU RAM and CPU Intel Xeon Gold 6226R CPU @ 2.90GHz with 15 cores, 
thus increasing the batch size from 16 to 64.
In this case, we directly drop the merging block pretrained weights.
This behavior suggests that the merging block weights (like the expander ones) are very important only during the pretraining phase. In other words, the merging block in VICReg helps the RGB and flow branches to learn the correct features but, with the increase of the batch size, the role of the merging block weights remains fundamental in VICReg while losing its relevance in the target task.  
The final results shown in Table~\ref{tab:dataset_vicreg_results} meet our expectations. We reach an \textbf{accuracy and F1-score of 86.5\% (+0.63\%)} which is a solid improvement compared to the baseline model. 
The confusion matrix shown in Fig.~\ref{fig:roc_and_confusion} (left) indicates that the model reaches \textbf{88\% (+2.5\%) in TNR} and \textbf{85\% (-1.25\%) in TPR}. While a decrease in the TPR rate is acceptable, the increase in TNR is well received. In fact, from an application point of view, the TNR is an important metric that avoids stressing out the user with excessive false positives, given that the number of non-violence samples would be greater than the violent ones in a real case scenario. Furthermore, by reducing false positives,  we can mitigate the risk of wrongful actions that could be taken against individuals who are mistakenly identified as violent.
For the sake of completeness, we show also the ROC curve in Fig.~\ref{fig:roc_and_confusion} (right). 
The \textbf{AUC reaches a value of 0.924} which is similar to the baseline model, demonstrating a good capability to distinguish between classes.
Although JOSENet performs slightly worse than the state of the art for RWF-2000 \cite{cheng_rwf-2000_2020}, it clearly represents a more efficient and faster solution thanks to a four-time smaller segment length (16 instead of 64) and a smaller FPS required (7.5 instead of 12.8). This proves that the proposed framework features an excellent compromise between performance and efficiency. 

\begin{table*}[t]
\centering
\caption{The evolution of the results of our JOSENet framework. The first row shows the results of the baseline model. The final results are shown in the last row.}
\vspace{0.1cm}
\label{tab:performance_evolution}
\begin{tabular}{ |c|c|c|c|c|c|c| } 
\hline
Dataset & Accuracy [\%] & F1 [\%] & TNR [\%] & TPR [\%] & AUC \\
\hline
No-pretraining & 85.87 &85.87 & 85.5 & 86.25 & 0.924 \\
R-UCF-Crime & \textbf{86.75} & \textbf{86.74} & 84.5 & \textbf{89} & 0.918 \\
UCF-Crime & 86.5 & 86.5 & \textbf{88} & 85 & \textbf{0.924}\\
\hline
\end{tabular}
\end{table*}

We outlined the JOSENet performance evolution in Table~\ref{tab:performance_evolution}. The first model does not involve any pretraining: the primary task has been addressed directly by the FGN trained on the RWF-2000 training set. Then, we implemented several non-regularized SSL strategies that produced slightly improved results, compared to the baseline. For this reason, we used VICReg as an auxiliary SSL model in our JOSENet framework. With the aim of selecting the best pretraining dataset for JOSENet we have seen that R-UCF-Crime obtained the higher results (see the second row of Table~\ref{tab:performance_evolution}). Finally, we pretrained the auxiliary SSL VICReg model on the entire UCF-Crime, reaching the final results for JOSENet, shown in the last row of Table~\ref{tab:performance_evolution}.

%

\begin{table*}[t]
\centering
\caption{
Representations from InfoNCE, UberNCE, CoCLR and JOSENet are evaluated on the target task, with a pretraining obtained on a R-UCF-101. The \xmark \ indicates the random initialization of the branch during target training. In the last rows, we simultaneously initialize both the RGB and flow blocks of the FGN model.}
\vspace{0.1cm}
\label{tab:sota_comparison}
\begin{tabular}{ |c|c|c|c|c|c|c|c|c| } 
\hline
Method & RGB & Flow & Accuracy [\%] & F1 [\%] & TNR [\%] & TPR [\%] & AUC \\
\hline
InfoNCE \cite{infonce_2019}& \cmark & \xmark & 85.12 & 85.12 & 84 & 86 & 0.906 \\
UberNCE \cite{co_clr_2020} & \cmark & \xmark & 84.5 & 84.5 & 77 & \textbf{92} & 0.911\\
CoCLR \cite{co_clr_2020} & \cmark & \xmark & 84.25 & 84.25 & 84 & 84 & 90.86\\
JOSENet (ours) & \cmark & \xmark & 84.62  & 84.62 & 85.75 & 83.5 & 0.917\\
\hline
InfoNCE \cite{infonce_2019} & \xmark & \cmark & 85.12 & 85.12 & 87 & 83 & 0.907\\
UberNCE \cite{co_clr_2020} & \xmark & \cmark & 84 & 83.99 & 86 & 82 & 0.905\\
CoCLR \cite{co_clr_2020} & \xmark & \cmark & 84.25 & 84.25 & 84 & 84 & 0.908\\
JOSENet (ours) & \xmark & \cmark & 84.49 & 84.49 & 82.75 & 86.75 & 0.900\\
\hline
InfoNCE \cite{infonce_2019} & \cmark & \cmark & 83.62 & 83.59 & 89 & 79 & 0.897 \\
UberNCE \cite{co_clr_2020} & \cmark & \cmark & 83.87 & 83.80 & \textbf{91} & 77 & 0.895 \\
CoCLR \cite{co_clr_2020} & \cmark & \cmark & 83.5 & 83.44 & 90 & 78 & 0.898\\
JOSENet (ours) & \cmark & \cmark & \textbf{86.4} & \textbf{86.4} & 85.75 & 87 & \textbf{0.922}\\

\hline
\end{tabular}
\end{table*}

%
\subsection{SSL State-of-the-Art Comparison}
In this Section, we compare JOSENet with previous SOTA self-supervised approaches: InfoNCE \cite{infonce_2019}, UberNCE \cite{co_clr_2020} and CoCLR \cite{co_clr_2020}. 
We decided to take as reference the results obtained on R-UCF-101 with JOSENet. For a fair comparison, we pretrain on the same  dataset either RGB or flow blocks using the SOTA methods by scaling down accordingly some of their hyperparameters. 
Then, as usual, the obtained weights are used as pretraining for the target task. The results are shown in Table~\ref{tab:sota_comparison}.
We can observe that InfoNCE surpasses both the accuracy and F1-score performance of JOSENet and all the other methods when only the RGB or the flow blocks is pretrained. However, the AUC results show that UberNCE is still a valid alternative to the instance-based SSL in such a situation. 
As expected, when two pre-trained branches are used, CoCLR has better AUC performance compared to all the other SOTA methods thanks mainly to the co-training scheme. Nevertheless, a huge drop in performance happens in the SOTA methods compared to JOSENet, which obtains the best results on this setting: +1.28\% accuracy and F1-score, with +0.011 AUC from the second best method. These results can be explained by the fact that all the SOTA methods do not take into consideration the merging block of the FGN while JOSENet is able to exploit that part of the architecture during pretraining, thus producing better embeddings for the target model.
\subsection{Generalizing JOSENet to Action Recognition Tasks}
As an additional test, we use our JOSENet framework pretrained with VICReg on the UCF-Crime dataset as a starting point for the action recognition task. In particular, we fine-tune the FGN on two different benchmark datasets for action recognition: HMDB51 and UCF101.
In order to avoid any overfitting of the network on dynamic actions, the only modification that we apply is the removal of the RoI extraction for the RGB frames. The entire training procedure and the architecture remain unchanged compared to the one described in the previous sections.
\begin{table}[t]
\centering
\caption{Action recognition results of our FGN pretrained with VICReg on UCF-Crime. The last two columns represent respectively the Top-1 and Top-5 accuracy computed on fold 1 for both datasets.}
\label{tab:action_recognition_results}
\begin{tabular}{ |c|c|c|c| } 
\hline
Dataset & VICReg Pretraining & Top-1 [\%]& Top-5 [\%] \\
\hline
HMDB51& \xmark &20 & 49.03 \\
HMDB51& \cmark &\textbf{24.62} & \textbf{55.03} \\
\hline
UCF101& \xmark &43.18 & 70.12 \\
UCF101& \cmark &\textbf{48.1} & \textbf{73.24} \\
\hline
\end{tabular}
\end{table}

The results in Table~\ref{tab:action_recognition_results} show that our pretraining based on the VICReg seems to be effective also for a more generic action recognition task. Indeed, in both the datasets we obtain a boost in performances of about 3-6\% of accuracy.
We want to highlight that our architecture trained from scratch surpasses by 1-3\% Top-1 accuracy of the Resnet3D-18 \cite{hara_can_2018}, which requires a huge computational power (33.3M parameters) compared to the FGN network (272,690 parameters).
Thus, we strongly believe our JOSENet framework could obtain near state-of-the-art performance by exploiting a pretraining obtained on a larger dataset.

%
%
%
%
%
\section{Ablation Studies}
\label{sec:ablation}
\subsection{Siamese Architecture} 
To evaluate the effectiveness of a simplified Siamese architecture for pretraining, we pretrain either the RGB or flow branches using shared weights. In both scenarios, the branches are pretrained on the HMDB51 dataset. During the training phase of the target model, the block that was not pretrained is initialized with random weights. Additionally, we conduct an experiment where, to circumvent random initialization, the pretrained blocks are simultaneously utilized during the primary model training. The results are shown in Table~\ref{tab:vicreg_siamese_results}. 
All these approaches appear to be inefficient, indicating that to enhance the embeddings of both the RGB and flow branches, JOSENet must leverage the complementary information provided by both RGB and flow views during the pretraining phase.
\begin{table}[t]
\centering
\caption{Results of the target task achieved through pretraining with the VICReg method on HMDB51 dataset, utilizing two Siamese branches, either RGB or flow. 
The \xmark \ indicates the random initialization of the branch during target training. 
In the last row, we simultaneously used the previously pretrained Siamese models on the primary model.}
\label{tab:vicreg_siamese_results}
\begin{tabular}{ |c|c|c|c|c|c|c| } 
\hline
RGB & Flow & Accuracy & F1 [\%] & TNR [\%] & TPR [\%] & AUC \\
\hline
\cmark & \xmark & 84.62 & 84.62 & 85.75 & 83.5 & 0.917 \\
\xmark & \cmark & 84.5 & 84.49 & 82.07 & 87.03 & 0.918 \\
\cmark & \cmark & 84.37 & 84.37  & 82.75 & 86.75 & 0.900 \\
\hline
\end{tabular}
\end{table}
\begin{table*}[t]
\centering
\caption{Results obtained on the target task by pretraining the VICReg on HMDB51 using different configurations. In particular, we test our model by including or removing the zoom crop (ZC) augmentation strategy and/or the temporal max-pooling (TMP) in the merging block.}
\vspace{0.1cm}
\label{tab:vicreg_road_to}
\begin{tabular}{ |c|c|c|c|c|c|c| } 
\hline
ZC & TMP & Accuracy [\%] & F1 [\%]& TNR [\%] & TPR [\%] & AUC \\
\hline
\cmark & \xmark & \textbf{85.37} & \textbf{85.37} & 84.01 & \textbf{86.75} & \textbf{0.924} \\
\cmark & \cmark & 82.5 & 82.49 & 84.75 & 80.25 & 0.8927 \\
\xmark & \xmark & 85.25 & 85.22 & \textbf{90.15} & 81.57 & 0.9159 \\
\hline
\end{tabular}
\end{table*}

%
\subsection{Augmentation Strategies}
The results on different architecture and augmentation strategies are shown in Table~\ref{tab:vicreg_road_to}.
We first pretrain our model by using the zoom crop strategy. At the same time, we try to avoid an excessive reduction in the dimensionality of the expander input by removing the temporal max-pooling in the merging block.
While the results on the accuracy are slightly lower than our baseline, we obtain an increase in TPR by +0.50\% and a slight increase in AUC suggesting a feasible model configuration.
To find a confirmation of this approach, using the zoom crop strategy, we apply the temporal pooling in the merging block, obtaining on the target task a very low value for most of the evaluation metrics used. This test shows that some issues occur when there is a very small bottleneck between the encoder and the expander. In fact, the input expander dimensionality is reduced from 1024 to 128.
Successively, the network is pretrained with the random crop \cite{bardes_vicreg_2022} while avoiding the temporal pooling in the merging block. The results are not optimal 
and suggest that zoom crop augmentation is crucial for the network to extrapolate the most useful features. 
\section{Conclusion}
\label{sec:conclusion}
In this work, we introduced JOSENet, a novel regularized SSL framework involving a modified VICReg for a two-stream video architecture. The proposed framework is able to tackle effectively the violence detection task, a challenging research topic in computer vision.
The proposed JOSENet framework has proven to achieve high performance for violence detection, while maintaining solid generalization capability, as it is able also to detect non-violent actions. These results are the basis for product stability from an application deployment perspective. Furthermore, our experiments showcase that JOSENet outperforms state-of-the-art SSL methods in the domain of violence detection, underscoring its superior performance in this task. Lastly, we assess the robustness and generality of the framework by deploying it in generic action recognition tasks.
In the future, it would be interesting to focus on possible limitations, e.g., avoiding any possible bias in unfair prediction, reducing the risk of false negatives, assessing the robustness against real-world issues on data (e.g., occlusions, light conditions), and improving the efficiency of the optical flow branch.

%
%
%
%
%
\section*{Reproducibility}
The source code and instructions for reproducing our experimental results are available at {\href{https://github.com/ispamm/JOSENet}{github.com/ispamm/JOSENet}}.

All the data used in this paper are publicly available. The UCF-Crime dataset \cite{sultani-ucf-crime-2019} is composed of 128 hours of real-world surveillance videos and it is available at \url{https://www.crcv.ucf.edu/data/UCF101.php}. 
The RWF-2000 dataset \cite{cheng_rwf-2000_2020}, available at \url{https://github.com/mchengny/RWF2000-Video-Database-for-Violence-Detection}, is made of 2000 videos, 5 seconds long, and captured at 30 FPS by real-world surveillance cameras. 
The HMDB51 dataset \cite{hmdb51_2011} contains 51 classes spread over 6,766 clips and it is available at \url{https://serre-lab.clps.brown.edu/resource/hmdb-a-large-human-motion-database/}.
A larger dataset called UCF101 \cite{soomro_ucf101_2012} is available at \url{https://www.crcv.ucf.edu/data/UCF101.php}.

%
%
%
%
%
\bibliographystyle{ieeetr}
\bibliography{JOSENETrefs}

\end{document}